\documentclass[twoside]{article}

\usepackage[accepted]{aistats2019}

\usepackage{algpseudocode}
\usepackage{algorithm}
\usepackage{setspace}
\let\Algorithm\algorithm
\renewcommand\algorithm[1][]{\Algorithm[#1]\setstretch{1.1}}

\usepackage{scalerel}
\usepackage{amsthm}
\usepackage{amsmath}							
\usepackage{amssymb}
\usepackage{mathtools}
\usepackage{commath}

\usepackage{graphicx}	
\graphicspath{{../figs/}}
\usepackage{float}
\usepackage{afterpage}
\usepackage{caption}
\usepackage{subcaption}
\usepackage{layouts}

\usepackage[dvipsnames]{xcolor}

\usepackage{blindtext}

\usepackage[hidelinks]{hyperref}

\newcommand{\thetab}{{\boldsymbol{\theta}}}

\newcommand{\alphab}{{\boldsymbol{\alpha}}}

\newcommand{\mub}{\boldsymbol{\mu}}
\newcommand{\Sigmab}{\boldsymbol{\Sigma}}

\newcommand{\pnorm}{p}
\newcommand{\pnn}{\phi}

\newcommand{\pnoise}{p_{ \mathbf y}}

\newcommand{\q}[1]{q(\z \given{#1})}
\newcommand{\gradtheta}{\nabla_{\thetab}}

\renewcommand{\u}{{\mathbf u}}
\newcommand{\x}{{\mathbf x}}
\newcommand{\y}{{\mathbf y}}
\newcommand{\z}{{\mathbf z}}

\newcommand{\N}{\mathcal{N}}
\newcommand{\I}{{\mathbb I}}
\newcommand{\K}{{\mathbf K}}

\newcommand{\E}{\mathbb{E}}
\newcommand{\Ex}{\E_{\x}}
\newcommand{\Ey}{\E_{\y}}

\newcommand{\Evar}[1]{\E_{\z \sim \q{#1}}}

\newcommand{\JN}{J_{\scaleto{\mathrm{NCE}}{4 pt}}}
\newcommand{\JV}{J_{\scaleto{\mathrm{VNCE}}{4 pt}}}

\newcommand{\suppl}{supplementary material}
\newcommand{\classprob}{\kappa_{\x}}

\newcommand\given[1][]{\:#1\vert\:}

\DeclarePairedDelimiterX{\infdivx}[2]{(}{)}{%
#1\;\delimsize\|\;#2%
}
\newcommand{\infdiv}[1]{D_{#1}\infdivx}

\newcommand{\argmax}[1]{\underset{#1}{\operatorname{arg}\,\operatorname{max}}\;}

\DeclareMathOperator\erf{erf}
\DeclareMathOperator\erfcx{erfcx}

\newtheorem{theorem}{Theorem}
\newtheorem{corollary}{Corollary}
\newtheorem{lemma}{Lemma}
\theoremstyle{definition}
\newtheorem{definition}{Definition}

\usepackage[round]{natbib}

\newcommand{\comment}[1]{} 

\begin{document}

\twocolumn[
\aistatstitle{Variational Noise-Contrastive Estimation}
\aistatsauthor{Benjamin Rhodes \And Michael U. Gutmann}
\aistatsaddress{School of Informatics \\ University of Edinburgh\\  \href{mailto:ben.rhodes@ed.ac.uk}{ben.rhodes@ed.ac.uk} \And School of Informatics\\ University of Edinburgh\\ \href{mailto:michael.gutmann@ed.ac.uk}{michael.gutmann@ed.ac.uk}}]

\begin{abstract}
  Unnormalised latent variable models are a broad and flexible class
  of statistical models. However, learning their parameters from data
  is intractable, and few estimation techniques are currently
  available for such models. To increase the number of techniques in
  our arsenal, we propose variational noise-contrastive estimation
  (VNCE), building on NCE which is a method that only applies to
  unnormalised models. The core idea is to use a variational lower
  bound to the NCE objective function, which can be optimised in the
  same fashion as the evidence lower bound (ELBO) in standard
  variational inference (VI). We prove that VNCE can be used for both
  parameter estimation of unnormalised models and posterior inference
  of latent variables. The developed theory shows that VNCE has the
  same level of generality as standard VI, meaning that advances made
  there can be directly imported to the unnormalised setting. We
  validate VNCE on toy models and apply it to a realistic problem of
  estimating an undirected graphical model from incomplete data.
\end{abstract}

\section{Introduction}
Building flexible statistical models and estimating them is a core
task in unsupervised machine learning. For observed data $\{\x_1, \ldots,
\x_n\}$, parametric modelling involves specifying a family of
probability density functions (pdfs) $\{p(\x; \thetab)\}$ parametrised
by $\thetab$ that has the capacity to capture the structure in the
data. Two fundamental modelling techniques are (i) introducing latent
variables which serve as explanatory factors or model missing data; and
(ii) energy-based modelling which removes the constraint that each
member of the family has to integrate to one, rendering the model
unnormalised.

Both techniques are widely used. Latent variable models have generated
excellent results in an array of tasks, such as semi-supervised
modelling of image data \citep{kingma2014semi} and topic modelling of
text corpora \citep{hoffman2013stochastic}. In addition, many
real-world data sets are incomplete, and it is advantageous to model
the missing values probabilistically as latent variables
\citep[e.g.][]{jordan1999introduction, nazabal2018handling}. Energy-based models --- also known as
unnormalised models --- have led to several advances in
e.g.\ neural language modelling \citep{mnih2013learning}, multi-label
classification \citep{belanger2016structured} and unsupervised
representation learning \citep{oord2018representation}.

Despite their individual successes, there are few attempts in the
literature to combine the two types of models, a notable exception being deep Boltzmann
machines \citep{ruslan2009deep}. This is
primarily because learning the parameters of unnormalised latent
variable models is very difficult. For both types of models, evaluating
$p(\x; \thetab)$ becomes intractable, and thus the combined case is
doubly-intractable. For latent variable models, $p(\x; \thetab)$ is
only obtained after integrating out the latents $\z$
\begin{equation}
  p(\x; \thetab) = \int p(\x, \z; \thetab) \dif \z,
\end{equation}
whilst for unnormalised models $\pnn(\x; \thetab)$, we have
\begin{equation}
p(\x; \thetab) = \pnn(\x; \thetab) / Z(\thetab),
\end{equation}
where $Z(\thetab) = \int \pnn(\x; \thetab) \dif \x$ is the normalising
partition function. In both cases, the model $p(\x; \thetab)$ is
defined in terms of integrals that cannot be solved or easily
approximated. And without access to $p(\x; \thetab)$, we cannot
learn $\thetab$ by standard maximum likelihood estimation.

One potential solution is to make use of the following expression of
the gradient of the log-likelihood (for a data point $\x_i$)
\begin{align}
  &\E_{\z \sim p(\z \given \x_i; \thetab)} \left[ \gradtheta \log \pnn(\x_i, \z;\thetab) \right] -  \nonumber \\
  & \E_{\x, \z \sim p(\x, \z; \thetab)} \left[\gradtheta \log \pnn(\x, \z;\thetab) \right]
  \label{eq:log-lik-grad}
\end{align}
and to perform stochastic ascent on the log-likelihood. This requires
samples from $p(\x,\z;\thetab)$ and the posterior $p(\z \given \x_i ; \thetab)$ which, for some models, can be obtained by Markov
chain Monte Carlo.

However, this approach is not always practical, or even feasible, and so
more specialised methods are being used for efficient parameter
estimation. To handle latent variables, variational inference
\citep{jordan1999introduction} is a commonly used, powerful technique
involving the maximisation of a tractable lower bound to the
log-likelihood. For unnormalised models, specialised methods include
score matching \citep{hyvarinen2005estimation}, ratio matching
\citep{hyvarinen2007some}, contrastive divergence
\citep[CD,][]{hinton2006training}, persistent contrastive divergence
\citep{younes1998stochastic, tieleman2009using} and noise-contrastive
estimation \citep[NCE,][]{Gutmann2012a}.

There are thus multiple estimating methods for \emph{either} latent
variable models \emph{or} unnormalised models, but not for both and
there has been little work on combining methods from the two camps. To
our knowledge, the only combination available is (persistent) CD with
variational inference \citep{ruslan2009deep}. Whilst this combination
has worked well in the context of Boltzmann machines, it is unclear
how well these results generalise to other models. Given the
limited number of existing methods, it is important to have more
estimation techniques at our disposal. 

We here develop a novel variational theory for NCE that enables
parameter estimation of unnormalised, latent variable models. This
method, VNCE, maximises a variational lower bound to the NCE
objective. Just as with standard variational inference on the
log-likelihood, VNCE \emph{both} estimates the model parameters and
yields a posterior distribution over latent variables. For parameter
estimation, we prove that VNCE is, in a sense, equivalent to NCE and
is theoretically well grounded. For approximate inference, we prove
that VNCE minimises a f-divergence between the true and approximate
posterior. We further prove that with increased use of computational
resources, we can recover standard variational inference by pushing
this f-divergence towards the usual Kullback-Leibler (KL) divergence.

Given the applicability of NCE in a wide range of domains such as
natural language processing \citep[NLP, e.g.][]{mikolov2013distributed},
recommendation systems \citep{huang2015neural}, policy transfer
learning \citep{zhang2016collective} etc.\ the ability to now
incorporate latent variables into such models is an important
advance. For instance, in NLP, latent variable models, such as topic
models, are in widespread use \citep{kim2018tutorial}. Moreover,
missing data is a ubiquitous problem across many domains and, as we
show experimentally in Section \ref{sec:trunc-norm}, modelling the
missing values probabilistically with VNCE offers signficant gains
compared to using NCE with standard fixed-imputation strategies.

\section{Background}
\label{sec:nce}
Noise-contrastive estimation \citep[NCE,][]{Gutmann2012a} is a method
for estimating the parameters of unnormalised models $\pnn(\x;
\thetab)$. The idea is to convert the unsupervised estimation problem
into a supervised classification problem, by training a (non-linear)
logistic classifier to distinguish between the observed data $\{\x_1,
\ldots, \x_n\}$, and $m$ auxiliary samples $\{\textbf{y}_1, \ldots,
\textbf{y}_m\}$ that are drawn from a user-specified `noise'
distribution $\pnoise$.

Using the logistic loss, parameter estimation in NCE is done by
maximising the sample version of $\JN(\thetab)$, 
\begin{align}
  \JN(\thetab) &= 
   \Ex  \log h(\x; \thetab)
  + \nu \Ey \log ( 1 - h(\y; \thetab) ),
  \label{eq:J}
\end{align}
where $\nu = m/n$ and $h(\u; \thetab)$ depends on the unnormalised model
$\pnn(\u; \thetab)$ and the noise pdf $\pnoise(\u)$,\footnote{We use $\u$ as a dummy variable throughout the paper.}
\begin{align}
  h(\u; \thetab) & = \frac{\pnn(\u; \thetab)}{\pnn(\u; \thetab) + \nu \pnoise(\u)}.
\end{align}
Typically, the model is allowed to vary freely in scale which can
always be achieved by multiplying it by $\exp(-c)$, where $c$ is a
scaling parameter that we absorb into $\thetab$ and estimate along
with the other parameters.

\citet{Gutmann2012a} prove that the resulting estimator is consistent for
unnormalised models. They further show that NCE approaches the
performance of MLE as the ratio of noise to data samples $\nu$
increases \citep[for stronger results, see][]{barthelme2015poisson,
  RiouDurand2018}. For generalisations of NCE to other than the
logistic loss function, see \citep{Pihlaja2010,
  Gutmann2011b,barthelme2015poisson}.

The noise distribution affects the efficiency of the NCE
estimator. While simple distributions such as Gaussians or uniform
distributions often work well \citep{Mnih2012}, both
intuition and empirical results suggest that the noise samples should
be hard to distinguish from the data \citep{Gutmann2012a}. The choice of the noise
distribution becomes particularly important for high-dimensional data or when
the data is concentrated on a lower dimensional manifold. For recent
work on choosing the noise semi-automatically, see
\citet{ceylan2018conditional}.

While NCE avoids the computation of the intractable partition
function, it assumes that we have data available for all variables
$\x$ in the model. This means that NCE is, in general, not applicable
to latent variable models. It will only apply in the special case
where we can marginalise out the latent variables, as e.g.\ in mixture
models \citep{matsuda2018estimation}. The fact that NCE cannot handle
more general latent variable models is a major limitation that we
address in this paper.

\section{Variational noise-contrastive estimation}
\label{sec:vnce}
We here derive a variational lower bound on the NCE objective function, allowing us to estimate the parameters of unnormalised, latent variable models. We then provide theoretical guarantees for this novel type of variational inference.

\subsection{NCE lower bound}
\label{sec:nce lower bound}
We assume that we are given an unnormalised parametric model $\pnn(\x,
\z;\thetab)$ for the joint distribution of the observables $\x$ and
the latent variables $\z$ (some of which may correspond to missing
data). The unnnormalised pdf $\pnn(\x;\thetab)$ of the observables
$\x$ is then defined via the (typically intractable) integral
\begin{equation}
  \pnn(\x;\thetab) = \int \pnn(\x,\z;\thetab) \dif \z.
\end{equation}

The NCE objective function $\JN$ depends on $\pnn(\u; \thetab)$
through $\log h(\u; \thetab)$, which occurs in the first term of
$\JN$, and $\log (1-h(\u;\thetab))$, which occurs in the second
term of $\JN$. For the first term, we can write
\begin{align}
 \log  h(\x; \thetab)   &= \log \bigg(\frac{\int \pnn(\x, \z;\thetab) \dif \z}{\int \pnn(\x, \z;\thetab) \dif \z + \nu \pnoise(\x)}\bigg) \\
    &=   g(r(\x; \thetab)),
\end{align}
where we introduced the notation
\begin{align}
\hspace{-2ex}  g(r) &= -\log \bigg(1+\frac{\nu}{r} \bigg),&  r(\x;\thetab) &= \frac{\int \pnn(\x, \z;\thetab) \dif \z}{\pnoise(\x)}.
  \label{eq:r-def}
\end{align}
Importantly, $g$ is a concave function of $r$ (see the \suppl). Using importance sampling, we then rewrite $r$ as an expectation
\begin{equation}
r(\x;\thetab) = \Evar{\x} \bigg( \frac{\pnn(\x, \z;\thetab)}{\q{\x}\pnoise(\x)} \bigg) 
\end{equation}
and apply Jensen's inequality to obtain the bound
\begin{align}
 & g(r(\x; \thetab))
    \geq  \Evar{\x} g \bigg( \frac{\pnn(\x, \z;\thetab)}{\q{\x}\pnoise(\x)} \bigg) \\
  & \hspace{-4mm}   \geq  \Evar{\x} \log \bigg( \frac{\pnn(\x,\z; \thetab)}{\pnn(\x,\z; \thetab) + \nu\q{\x}\pnoise(\x)} \bigg),
\end{align}
where the second line is obtained by substituting in the definition of $g$ and then rearranging. We note that this result does not follow from standard variational inference on the log-likelihood, however it leverages the same mathematical trick of importance sampling combined with Jensen's inequality.

We now have a lower bound on the first, but not the second, term of
the NCE objective $\JN$. But this is actually sufficient: we can handle the
intractable integral in the second term with importance sampling,
re-using the \emph{same} variational distribution $q$ that we use in 
the first term. The final objective, which we call the VNCE
  objective, is then given by:
\begin{align}
  \label{eq:vnce objective}
\hspace{-4mm}  \JV(\thetab, q) 
  &= \Ex \Evar{\x} \log \bigg( \frac{\pnn(\x,\z; \thetab)}{\pnn(\x,\z; \thetab) + \nu\q{\x}\pnoise(\x)} \bigg) \nonumber \\
  & \hspace{-12mm} + \nu \Ey \log \bigg( \frac{\nu\pnoise(\y)}{\nu\pnoise(\y) + \E_{\z \sim q(\z \given \y)} \left[\frac{\pnn(\y,\z; \thetab)}{q(\z \given \y)} \right]}  \bigg). 
\end{align}
In practice, we optimise the sample version of this, replacing expectations with Monte Carlo averages.

By construction, we have that $\JN(\thetab) \ge \JV(\thetab, q)$ for
all $q$ and this bound is tight when the variational distribution
$q(\z \given \x)$ equals the true posterior $p(\z \given
\x;\thetab)$. Importantly, the true posterior is also the optimal
proposal distribution in the second term (see \suppl). Thus, we do not
need to blindly guess a good proposal distribution; we obtain one
automatically through maximising $\JV(\thetab, q)$ with respect to
$q$. Finally, we note that, just as with NCE, the user must specify
the noise distribution $\pnoise$. 

\subsection{Theoretical guarantees}
We here prove basic properties of VNCE and establish its connection to
NCE and standard variational inference. Below we simply state
the results; all proofs can be found in the \suppl.

Standard variational inference (VI) minimises the KL-divergence between the
approximate and true posterior. In contrast, we show that VNCE
minimises a different \emph{f-divergence} between the two posteriors.

\theoremstyle{definition}
\begin{definition}
An \emph{f-divergence} $\infdiv{f}{p}{q}$ between two probability
density functions $p$ and $q$, is defined as
\begin{equation}
    \infdiv{f}{p}{q} = \mathbb{E}_{u \sim q} \left[ f\left(\frac{p(u)}{q(u)}\right) \right],
\end{equation}
where $f$ is a convex function satisfying $f(1) = 0$.
\end{definition}
It follows from Jensen's inequality that f-divergences are
non-negative and obtain their minimum precisely when $p = q$. 
The KL divergence is an important example of an f-divergence, where $f(u) = u \log(u)$. 
\begin{lemma}
\label{lemma: diff J - J1}
The difference between the NCE and VNCE objective functions is equal to the expectation of an f-divergence between the true and approximate posterior. Specifically,\footnote{Throughout the following equations, parameters are moved into the subscript for compactness.}
\begin{equation}
        \JN(\thetab) - \JV(\thetab, q) = \Ex \left[ \infdiv{f_{\x}}{p_{\thetab}(\z \given \x)}{q(\z \given \x)} \right],
    \end{equation}
where
\begin{align}
    f_{\x}(u) &= \log(\classprob + (1 - \classprob)u^{-1}), & \classprob = \frac{\pnn_{\thetab}(\x)}{\pnn_{\thetab}(\x) + \nu \pnoise(\x)} .
\end{align}
Moreover, this f-divergence equals the difference of two KL-divergences
\begin{equation}
\label{eq: f-divergence is diff of two KLs}
\infdiv{KL}{q(\z \given \x)}{p_{\thetab}(\z \given \x)} - \infdiv{KL}{q(\z \given \x)}{m_{\thetab}(\z, \x)},
\end{equation}
where $m_{\thetab}(\z, \x) = \classprob p_{\thetab}(\z \given \x) + (1 - \classprob)q(\z \given \x)$ is a convex combination of the true and approximate posteriors.
\end{lemma}

The connection between standard VI and VNCE is
made explicit in \eqref{eq: f-divergence is diff of two KLs}, which
shows that VNCE not only minimises the standard KL, but also an
additional term: $- \infdiv{KL}{q(\z \given \x)}{m_{\thetab}(\z,
  \x)}$.

The following theorem shows that this additional term does not affect
the optimal non-parametric $q$, which is simply the true posterior
$p(\z \given \x ; \thetab)$. However, this additional KL term has an
impact when $q$ lies in a restricted parametric family not containing
the true posterior. Interestingly, by
increasing the ratio of noise to data, the extra KL term goes to zero
and we recover standard VI.

\begin{theorem}
\label{theorem: optimal q}
The VNCE lower bound is tight when $q$ equals the true posterior,
\begin{equation}
        \JN(\thetab) = \JV(\thetab, q) \hspace{2mm} \Leftrightarrow \hspace{2mm} q(\z \given \x) = p_{\thetab}(\z \given \x)
\end{equation}
and, as $\classprob = \pnn_{\thetab}(\x) / (\pnn_{\thetab}(\x) + \nu \pnoise(\x))  \rightarrow 0$, our f-divergence tends to the standard KL-divergence,
 \begin{align}
     \infdiv{f_{\x}}{p_{\thetab}(\z \given \x)}{q(\z \given \x)} \ \rightarrow \ \infdiv{KL}{q(\z \given \x)}{p_{\thetab}(\z \given \x)} .
 \end{align}
 In particular, as the ratio of noise to data, $\nu$, goes to infinity, we recover the standard KL-divergence.
\end{theorem}

The fundamental point of this theorem is that VNCE enables a valid
form of approximate inference. The fact that we recover the standard
KL-divergence as a limiting case is also of interest, and is in
agreement with a theoretical result for NCE, which states that as the
ratio $\nu$ tends to infinity, NCE is equivalent to maximum
likelihood (see Section \ref{sec:nce}).

A straightforward, but important, consequence of the foregoing theorem
is that joint maximisation of the VNCE objective $\JV$ with respect to the
variational distribution $q$ and model parameters $\thetab$ recovers the same solution as maximising the NCE objective with respect to $\thetab$.
\begin{theorem}{(Equivalence of VNCE and NCE)}
\label{theorem: equivalece of vnce and nce}
\begin{equation}
    \max_{\thetab} \JN(\thetab) = \max_{\thetab} \max_{q} \JV(\thetab, q)
\end{equation}
\end{theorem}
This theorem, which has its counterpart in standard VI, tells us that
VNCE is a valid form of parameter estimation. In particular, we could
maximise $ \JV(\thetab, q)$ by parametrising $q$ with parameters
$\alphab$, and jointly optimising with respect to both $\thetab$ and
$\alphab$. Alternatively, we may alternate between optimising
$\thetab$ and $\alphab$ as in variational EM. In either
case, we can use a score-function estimator
\citep{paisley2012variational, ranganath2014black, mnih2014neural} or
the reparametrisation trick \citep{kingma2013auto,
  rezende2014stochastic} to take derivatives with respect to
variational parameters $\alphab$.

In the special case that we know the true posterior over latents, we
no longer need to optimise $q$, and we obtain the (non-variational) EM
algorithm for VNCE. In the context of standard VI, the EM algorithm
can be very appealing because it never decreases the log-likelihood
\citep{dempster1977maximum}. We obtain an analogous result for VNCE,
shown in the following corollary.
\begin{corollary}{(EM algorithm for VNCE)}
\label{corollary:em for vnce}
    For any starting point $\thetab_0$, the optimisation procedure
    \begin{enumerate}
        \item (E-step) $q_k(\z \given \x) = p(\z \given \x ; \thetab_k) $ 
        \item (M-step) $\thetab_{k+1} = \argmax{\thetab} \JV(\thetab, q_k) $ 
        \item Unless converged, repeat steps 1 and 2
    \end{enumerate}
    never decreases the NCE objective function $\JN$, i.e.\ $\JN(\thetab_{k+1}) \geq \JN(\thetab_k)$ $\forall k \in \mathbb{N}$.
\end{corollary}
As is the case for standard EM, the above result does not hold if we
only take a `partial' E-step, by making $q$ close, but not exactly
equal, to $p(\z \given \x; \thetab)$ \citep{barber2012bayesian}. Thus,
any approach using a non-exact, variational $q$ will not have such
strong theoretical guarantees. However, the corollary still holds if
we take a partial M-step, increasing the value of $\JV(\thetab, q_k)$
by updating $\thetab$ through a few gradient steps.

\section{Validation and illustration of VNCE}
\label{sec:toy-models}
\subsection{Approximate inference with VNCE}
\label{sec: toy model for approx inference}
We here illustrate Theorem \ref{theorem: optimal q}, which justifies
the use of VNCE for approximate inference. For that purpose we
consider a simple normalised toy model $p(\x,\z)$ that has
2-dimensional latents and visibles,

\begin{align}
  p(\x, \z) & = p(\x \given \z) p(\z),&  p(\z) &= \N(\z ;0, \I),   \label{eq: toy posterior distribution1}\\
  p(\x \given \z) &= \N(\x ;\boldsymbol{\zeta}_{\z}, c^2 \I), & \boldsymbol{\zeta}_{\z} &= \begin{bmatrix} z_{1}z_{2} \\ z_{1}z_{2} \\ \end{bmatrix},   \label{eq: toy posterior distribution2}
\end{align}
where $c$ is fixed at $0.3$. Because $c$ is known, the model has no
parameters to estimate; we are solely interested in
approximating the posterior distribution $p(\z \given \x)$.

It does not appear possible to obtain a closed-form expression for
the exact posterior; instead we approximate it with $q(\z \given \x ;
\alphab) = \N(\z ; \ \mub(\x; \alphab), \Sigmab(\x; \alphab))$, where
$\Sigmab$ is a diagonal covariance matrix and the elements of $\mub$
and $\Sigmab$ are parametrised by a single 2-layer feed-forward neural
network--see the \suppl \; for details. This model can be viewed as a
simplified variational autoencoder \citep{kingma2013auto}, where the
decoder is not implemented with a neural network.

When applying VNCE, we consider two choices for the noise distribution
\begin{align}
    \pnoise^1(\y) &= \N(\y ; \ \Bar{\x}, \Bar{\Sigma}), & \pnoise^2(\y) = \N(\y ; \ 0, 30 \I),
\end{align}
where $\Bar{\x}$ and $\Bar{\Sigma}$ are the empirical mean and
covariance, respectively. The first choice is a `good' noise, that
matches the data well, whilst the second is a `bad' noise, poorly
matching the data. Figure \ref{fig: marginals for posterior
  visualisation} visualises the latent variable model and the two noise distributions.

\begin{figure}[t!]
  \centering
  \includegraphics[scale=1.1]{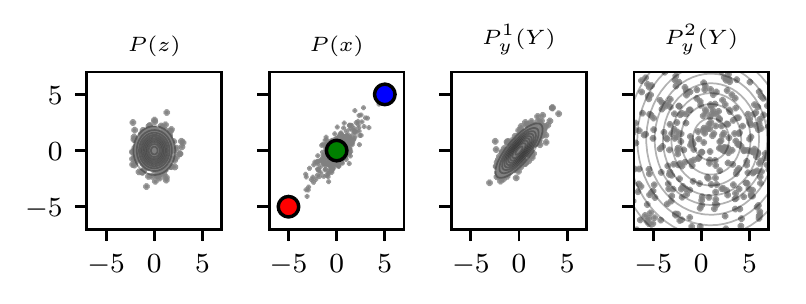}
 \caption{\label{fig: marginals for posterior visualisation} The two left-most plots are marginals of a latent variable model defined in \eqref{eq: toy posterior distribution1} and \eqref{eq: toy posterior distribution2}. The two right-most plots are noise distributions for VNCE.}
\end{figure}
\begin{figure}[t!]
  \centering
  \includegraphics[scale=0.8]{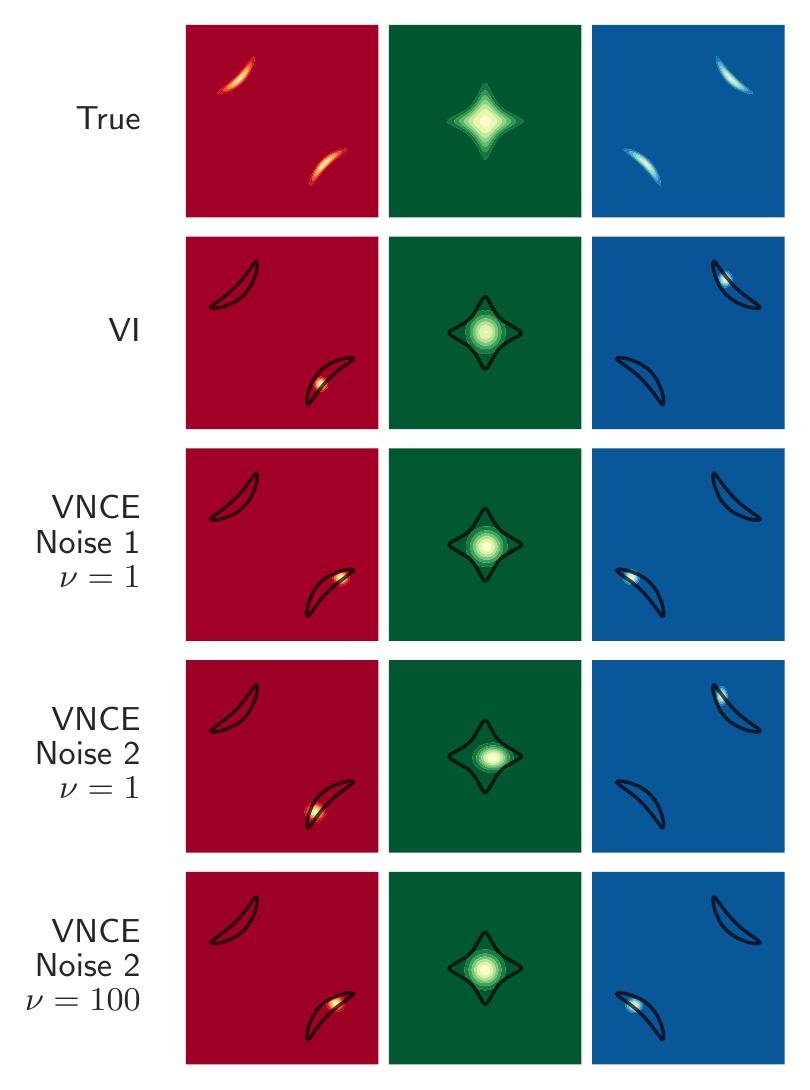}
  \captionof{figure}{ \label{fig:density plots of posteriors} Density plots of true and approximate posteriors for the 2D toy model defined in \eqref{eq: toy posterior distribution1} and \eqref{eq: toy posterior distribution2}. The colour-coded columns correspond to the landmark $\x$ points in the second plot of Figure \ref{fig: marginals for posterior visualisation}, which we condition on when computing posteriors.}
\end{figure}
Figure \ref{fig:density plots of posteriors} shows various posteriors over the latent space, conditioning on three colour-coded landmark $\x$ points marked in Figure \ref{fig: marginals for posterior visualisation}. The first two rows show the true posterior, calculated with numerical integration, and the approximate posterior learned using standard VI. Approximate posteriors learned with VNCE are shown in the last three rows.
The approximate posteriors learned with VNCE are similar to those
learned with standard VI when either the noise is a good match to the
data (row 3), or when $\nu$ is large (final row). In particular, the
VNCE posteriors show the same low-variance, mode-seeking
behaviour.

These connections between VNCE and standard VI are in line with
Theorem \ref{theorem: optimal q} which states that as the ratio
$\pnn_{\thetab}(\x) / (\pnn_{\thetab}(\x) + \nu \pnoise(\x))$ tends to
$0$, VNCE minimises an f-divergence that approaches the standard
KL. This ratio becomes closer to zero precisely when the noise assigns
a higher probability to the data or when $\nu$ is large. Conversely,
when the noise is `bad' and $\nu$ in insufficiently large (penultimate
row) VNCE produces approximate posteriors that are slightly distorted
in comparison to standard VI.

Theorem \ref{theorem: optimal q} also states that the optimal $q$
obtained with VNCE is the true posterior. In this setting, it is not
possible for $q$ to exactly recover the true posterior, since we have
restricted $q$ to be Gaussian with no correlation structure. Still, we
see that the approximate posteriors of both VI and VNCE are
reasonable fits to the true posteriors, modulo parametric
restrictions.

\begin{figure}[!t]
\centering
\includegraphics[scale=0.7]{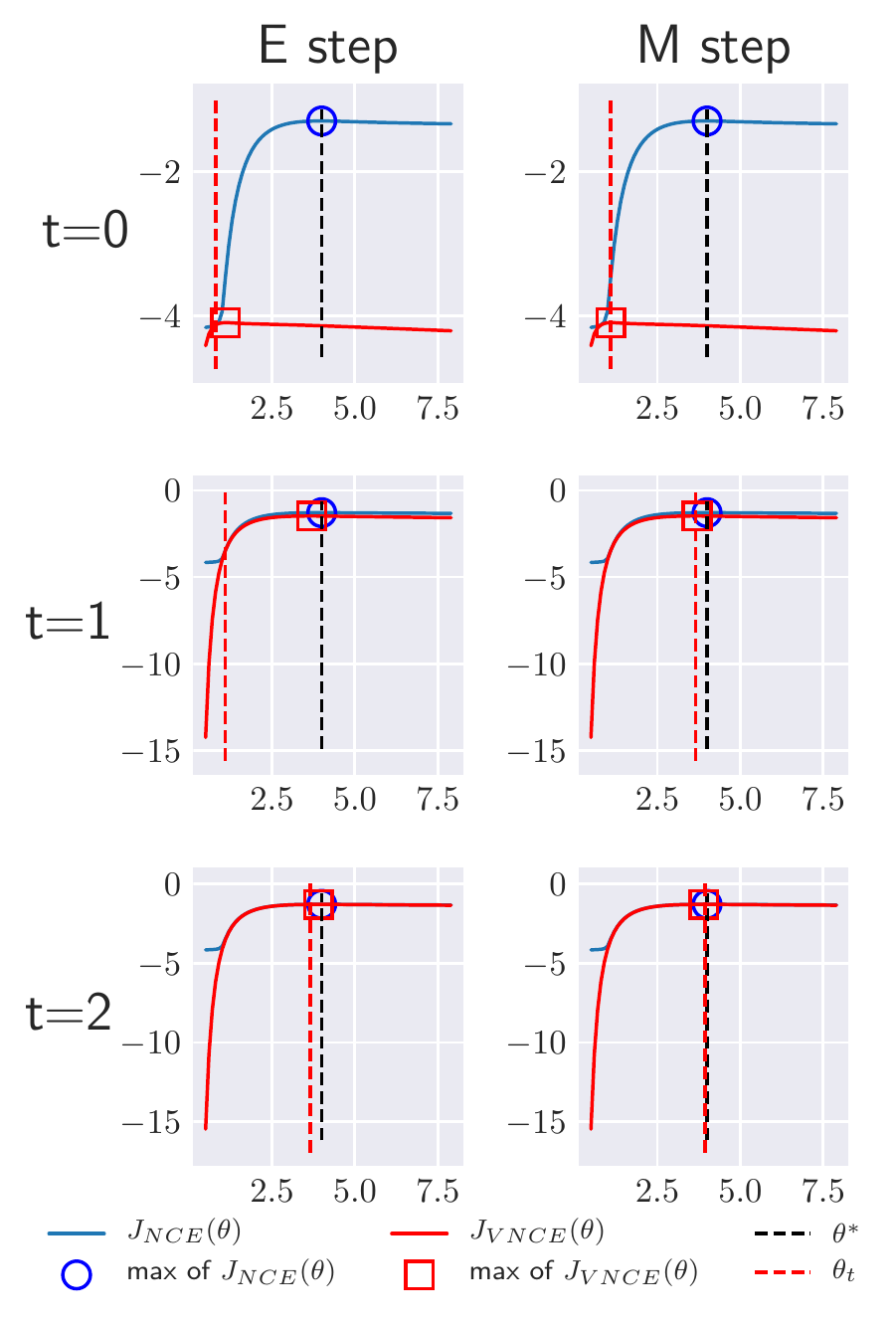}
\caption[The EM algorithm for VNCE, applied to a normalised Mixture of
  Gaussians]{\label{fig: normalised mog em} EM-type algorithm for VNCE. The figure reads row-by-row, from left to right. 
  In the E-step, we set $q(z|x)$ equal to the true posterior $p(z|x; \theta_t)$, making the VNCE objective tight at $\theta_t$. In the M-step we optimise $\theta$ using the VNCE objective, and hence the red dashed line shifts to the centre of the red square.}
\end{figure} 

\subsection{Parameter estimation with VNCE}

The following simulations illustrate Theorem \ref{theorem: equivalece of vnce and nce}, which states that VNCE and NCE have the same maximum. We consider both a normalised and unnormalised mixture of two Gaussians (MoG).

\paragraph{Normalised mixture of Gaussians}
The model is given by
\begin{equation}
\pnorm(x , z; \theta) = \frac{z}{2} \mathcal{N}(x; 0, \sigma_1^2) + \frac{(1-z)}{2} \mathcal{N}(x; 0, \theta^2),
\label{eq:mog standard}
\end{equation}
with $z \in\{0,1\}$ and $x \in \mathbb{R}$. We assume that the
variance of the first component, $\sigma_1^2$, is known, and we
estimate the value of $\theta$. For a simple experiment, we set
$\sigma_1 = 1$ and let $\theta^* = 4$ be the true value of
$\theta$. We set the noise distribution to be $\pnoise(y) =
\mathcal{N}(y; 0, {\theta^*}^2)$. For the variational distribution, we
can use the true posterior of the model,
\begin{align}
\label{eq:normalised mog posterior}
    p(z=0 \ | \ x; \theta) = \frac{1}{1 + \frac{\theta}{\sigma_1}  \exp ( \frac{- x^2}{2} ( \frac{1}{\sigma_1^2} - \frac{1}{\theta^2} ) ) },
\end{align}
enabling us to apply the EM type algorithm presented in Corollary \ref{corollary:em for vnce}.

Figure \ref{fig: normalised mog em} illustrates the results with plots
of the NCE and VNCE objectives obtained after each E-step and M-step
during learning. It is clear from the figure that the value of the NCE
objective at the current parameter (red-dashed line) never decreases,
in accordance with Corollary \ref{corollary:em for vnce}. Moreover,
the figure validates Theorem \ref{theorem: equivalece of vnce and
  nce}, which states that the maximum of the VNCE objective with
respect to $\theta$ and $q$ equals the maximum of the NCE objective
with respect to $\theta$. We see this from the overlap of the blue
circle (maximum of NCE) and the red square (maximum of VNCE) in the
final plot (bottom-right).

\begin{figure*}[!t]
      \centering
      \includegraphics[width=0.4\textwidth]{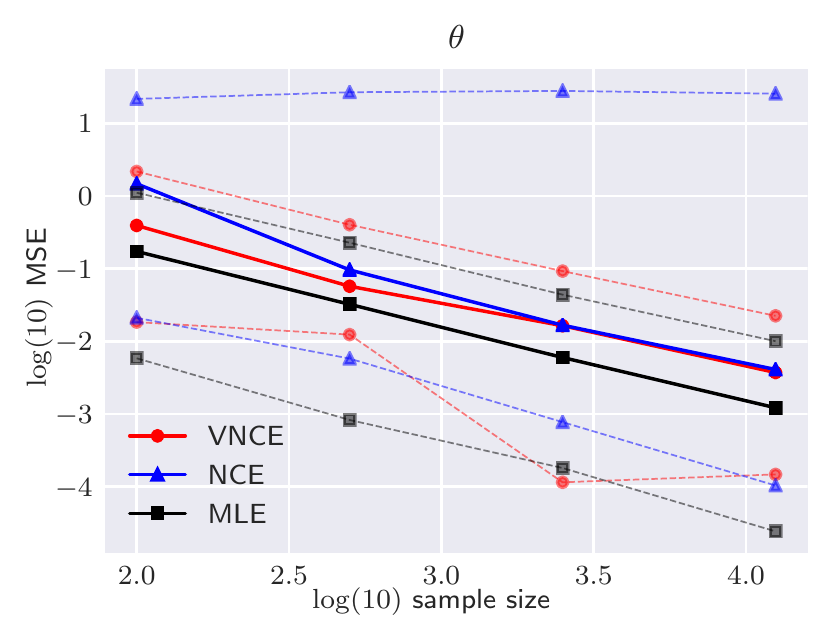}\hspace{3ex}
      \includegraphics[width=0.4\textwidth]{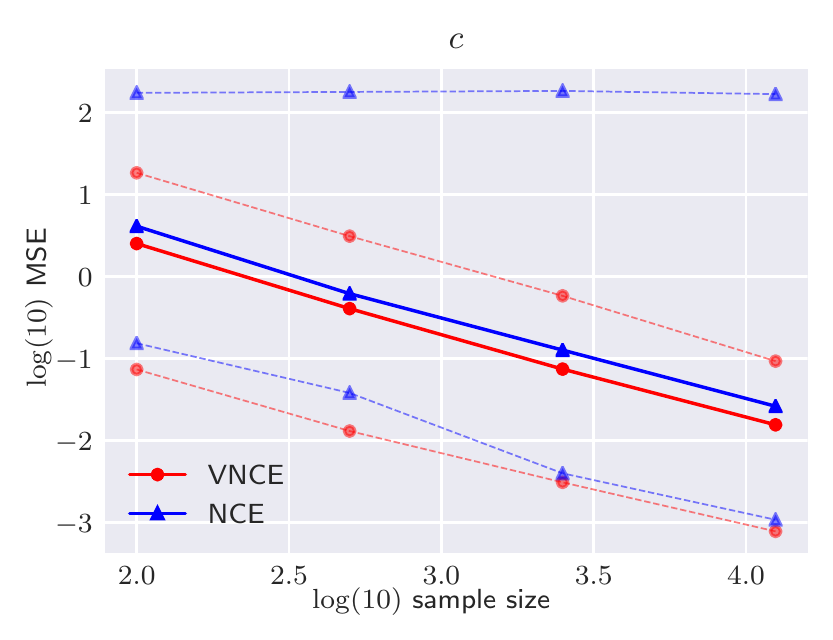}
    \caption[Population analysis for unnormalised mixture of Gaussian model]{ \label{fig:mse-mog} Log sample size vs.\ log mean-squared error for the standard deviation and scaling parameter of 500 different unnormalised MoG models. Central lines show median MSEs over 500 runs, whilst dashed lines mark the 1st and 9th deciles. The negative slope of the red line in both plots is evidence of the consistency of VNCE.
}
\end{figure*}
\paragraph{Unnormalised mixture of Gaussians}
    
An unnormalised version of the MoG model is given by
\begin{equation}
\phi(x, z; \theta, c) = e^{-c} \left( z e^{-\frac{x^2}{2 \sigma_1^2}} + (1-z) e^{-\frac{x^2}{2 \theta^2}} \right)
\label{eq:mog unnormalised}
\end{equation}
where $c$ is a scaling parameter. 

Whilst we could proceed as before, using an EM algorithm with the true posterior, we will not have access to such a posterior for more complex models. Thus, we test the performance of VNCE when using an approximate variational distribution $q$, given by
\begin{equation}
    q(z=0 \ | \ x; \boldsymbol{w}) = \frac{1}{1 + \exp(w_0 + w_1x + w_2x^2)},
\end{equation}
where $\boldsymbol{w} = (w_0, w_1, w_2)^\top$ are the variational parameters. This $q$ family contains the true posterior.

We test the accuracy of VNCE for parameter estimation using a
500-run population analysis over multiple sample sizes. NCE and maximum
likelihood estimation (MLE) serve as baseline methods (after
normalisation and/or summing over latent variables). For both NCE and
VNCE, we used $\nu=1$ and the same Gaussian noise distribution as for
the normalised MoG (for more details, see the \suppl).

Figure \ref{fig:mse-mog} shows the mean square error (MSE) $\mathbb{E}
||\theta - \theta^*||^2$ for VNCE, NCE and MLE. The left plot demonstrates that the estimation accuracy of VNCE increases with sample size, and is comparable to that of
NCE. This gives evidence of the consistency of
VNCE. Interestingly, NCE was much more prone to falling into local
optima, despite multiple random initialisations, as shown by the blue
upper dashed line.

\section{Graphical model structure learning from incomplete data}
\label{sec:trunc-norm}
We consider an important use-case of VNCE: the training of
unnormalised models from incomplete data, treating missing values as
latent variables. Specifically, we use VNCE to estimate the parameters
of an undirected graphical model from incomplete data. This
application is motivated by \citet{lin2016estimation}, who used
(non-negative) score matching \citep{hyvarinen2007some} for
estimation. Unfortunately, latent variables cannot be handled within
the score matching framework and so the missing values were either
discarded or set to zero.

\subsection{Model specification}
\label{sec:trunc norm model}
The undirected graphical model is a truncated Gaussian given by
\begin{equation}
\label{eq:truncated gaussian}
    \pnn(\x; \K, c) = \exp \left( -\frac{1}{2} \x^\top \K \x - c \right) \I(\x \in A) ,
\end{equation}
where $A$ is the support of $\pnn$, which equals $[0, +\infty]^d$ in
our experiments, and $c$ is a scaling parameter. The partition
function of $\pnn$ is intractable to compute, except in very low
dimensions \citep{horrace2005some}, rendering the model unnormalised.

The model in \eqref{eq:truncated gaussian} defines an undirected graph
where the variables correspond to nodes and where there is an edge
between the nodes of $x_i$ and $x_j$ whenever the $(i,j)$-th element
of $\K$ is non-zero. In such graphs, a missing edge between $x_i$ and $x_j$ means
that they are conditionally independent given the remaining variables
\citep[see e.g.][]{Koller2009}.

We split each data point $\x_i = (\x_i^{o}, \x_i^{m})$
into its observed and missing components. We treat the (potentially
empty) set of missing values $\x_i^{m}$ as latent variables, i.e.\ they
correspond to the $\z$ variables used before. The true posterior over
these missing variables, whilst also a truncated normal, is generally
intractable to compute \citep{horrace2005some}. We therefore use a
log-normal variational family to approximate it.

A subtle but important technical point is that there are
$2^d -1$ non-trivial patterns of missingness that can occur in the
data, and so we need a variational posterior for each possible
pattern. We achieve this by parametrising a \emph{joint} lognormal
distribution over all dimensions, since all of its conditionals are
computable in closed-form.

Similarly, we require noise samples, $\y_i=(\y_i^{o}, \y_i^{m})$, that
have the same pattern of missingness as the $\x_i$. In order to compute
the probability of $\y_i^{o}$, we need a joint noise distribution for
which we can compute all marginals. We achieve this by using a
fully-factorised product of truncated normals. The parameters of each
univariate truncated normal is estimated from the observed data for
that dimension (see \suppl).

\subsection{Simulations}
\label{sec:trunc norm simulations}

\begin{figure*}[t!]
    \centering
    \includegraphics[width = 0.425\textwidth]{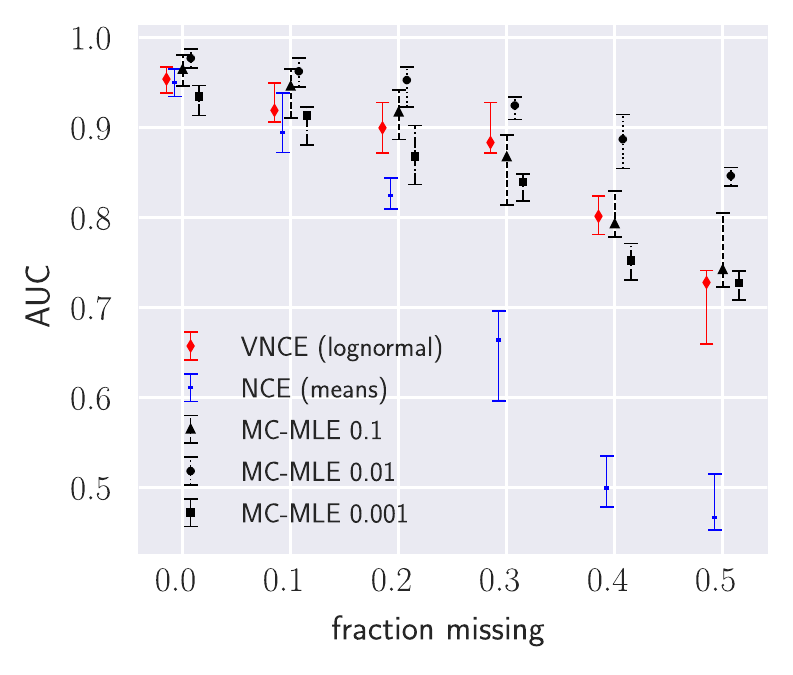} \hspace{3ex}
    \includegraphics[width = 0.425\textwidth ]{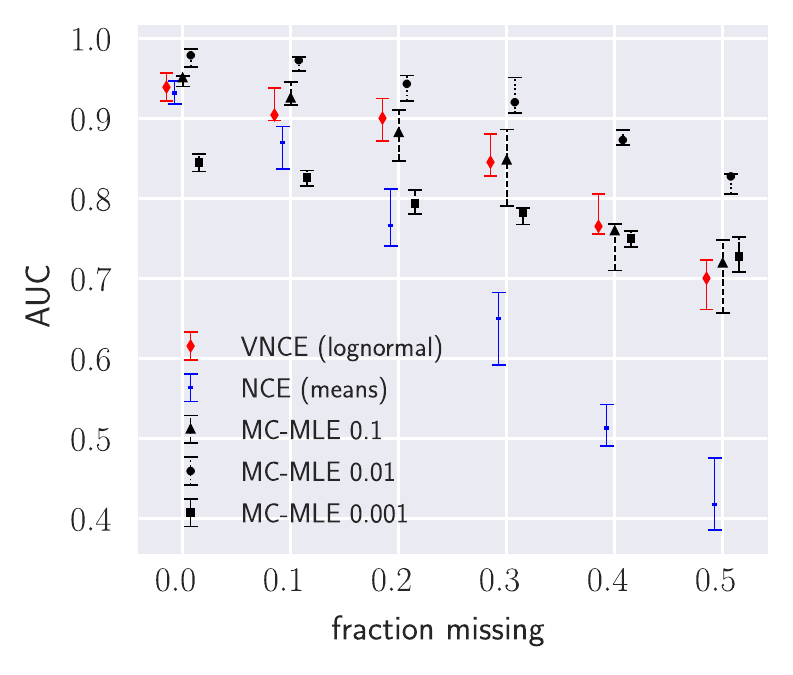}
    \caption[]{\label{fig:20d_auc}Left: ring-graph. Right: hub. Area under the ROC curve for increasing amounts of missing data. Larger AUC means better performance. Bars denote interquartile ranges for 10 runs, central markers medians.}
\end{figure*}

We consider two types of ground-truth graphs, and thus matrices
$\K$. The first is a ring-structured graph, where we obtain $\K$ from
an initial matrix of all-zeros by first sampling each element of the
superdiagonal from $\mathcal{U}(0.3, 0.5)$, as well as the top-right
hand corner, and then symmetrising. The second type of graph is an
augmented version of the ring-graph, where we have added `hubs',
i.e.\ nodes with a high degree. We randomly select $1/10$ nodes to be
connected to $1/4$ of all other nodes. 
In both cases, we set the diagonal elements to
a common positive number that ensures $\K$ is diagonally dominant.

We simulate 10 datasets of $n=1000$ samples with $d=20$ dimensions
using the Gibbs sampler from the \texttt{tmvtnorm} package in
\texttt{R} with a burnin period of 100 samples and thinning factor of
10. For each dataset, we generate six more, by discarding a percentage
$p$ of the $n \times d$ values at random, where $p$ ranges from $0\%$
to $50\%$ in increments of $10\%$.

We compare three methods: (i) VNCE, (ii) NCE with missing values
filled-in with the observed mean for that dimension, and (iii)
stochastic gradient ascent on the log-likelihood using the gradient in
\eqref{eq:log-lik-grad}, with the expectations approximated
via Monte Carlo sampling (MC-MLE). While VNCE and NCE were optimised with
a standard optimiser (BFGS), for MC-MLE, one has to manually select suitable
step-sizes for gradient ascent (for more details, see the \suppl).

For each data set and method, we can extract a learned graph from the
estimated $\K$ by applying a threshold. If an element of $\K$ is less
than the threshold, the corresponding edge is not included in the
graph. For various thresholds, we then compute a true-positive rate as
the percentage of ground-truth edges we correctly identify. Similarly,
we compute a false-positive rate. Jointly plotting the two rates
yields an ROC curve, and we use the area under the ROC curve (AUC) as
the performance metric.

Figure \ref{fig:20d_auc} shows the
results for the ring-graph (left) and the graph with hubs (right). In
both cases, we observe significant, and increasing, performance gains
for VNCE over NCE (with mean imputation) as larger fractions of data are missing. This shows
that inference of the missing values from the observed ones improves
parameter estimation. The difference is particular stark when 40\% or
more data is missing, as NCE is hardly better than random guessing of
edges (which corresponds to an AUC of 0.5). 

With careful tuning of the learning rate, MC-MLE achieves the best
performance of all three methods. This makes sense, since MLE is the
gold-standard for parameter estimation. However, for other reasonable
(but non-optimal) learning rates, VNCE performs comparably. This is an
important finding for two reasons. Firstly, MC-MLE is not feasible for
many models due to the lack of an efficient sampler, and so it is
valuable to know that VNCE can serve as a reasonable
replacement. Secondly, when modelling actual data, it is not obvious
how to select the stepsize, and other hyperparameters, for MC-MLE, due
to the lack of a tractable objective function. VNCE, in contrast, has
a well-defined objective function that can be optimised with powerful
optimisers. Moreover, it can be used for cross-validation in
combination with regularisation.

\vspace{-0.9ex}
\section{Conclusions}
\label{sec:conclusions}
\vspace{-0.9ex}
We developed a new method for training unnormalised latent variable
models that makes variational inference possible within the noise-contrastive framework. This contribution addresses an important gap in
the literature, since few estimation methods exist for this
highly-flexible, yet doubly-intractable, class of models.

We proved that variational noise-contrastive estimation (VNCE) can be
used for both parameter estimation and posterior inference of latent
variables. The proposed VNCE framework has the same level of
generality as standard variational inference, meaning that advances
made there can be directly imported to the unnormalised setting.

The theoretical results were validated on toy models and we
demonstrated the effectiveness of VNCE on the realistic problem of
graphical model structure learning with incomplete data. By working
with a model for which sampling is tractable, we were able to assess
VNCE in its ability to reach the likelihood-based solution. We found
that VNCE performed well and that it is a promising option for
estimating more complex unnormalised latent variables models where the
sampling-based approaches become infeasible.

\subsubsection*{Acknowledgements}
We would like to thank Iain Murray for feedback on preliminary versions of this text. Benjamin Rhodes was supported in part by the EPSRC Centre for Doctoral Training in Data Science, funded by the UK Engineering and Physical Sciences Research Council (grant EP/L016427/1) and the University of Edinburgh.
This work has made use of the resources provided by the Edinburgh Compute and Data Facility (ECDF).

\bibliography{vnce_bib}

\clearpage
\begin{appendix}
\onecolumn
\section{Convexity result for NCE lower bound}
\label{appendix:convexity result}
For non-negative real numbers $a, b$ and $u$, the function 
\begin{equation}
    f(u) = \log(a + bu^{-1})
\end{equation}
is convex. We see this by differentiating $f$ twice:
    \begin{align}
        f'(u) = - \frac{b}{au^2 + bu} \hspace{7mm} f''(u) = \frac{b(2au + b)}{(au^2 + bu)^2},
    \end{align}
and observing that $f''(u) \geq 0$ since $a, b$ and $u$ are non-negative.

\section{Proof of Lemma \ref{lemma: diff J - J1}}
\label{appendix:proof of lemma}
Key to this proof is the following factorisation
\begin{equation}
    \pnn_{\thetab}(\x, \z) = \pnn_{\thetab}(\x) p_{\thetab}(\z \given \x) ,
\end{equation}
where the conditional distribution is normalised and the factorisation holds because the unnormalised distributions on either side of the equation have the same partition function
\begin{equation}
    \int \int \pnn_{\thetab}(\x, \z) \dif \z \dif \x = \int  \pnn_{ \thetab}(\x) \dif \x .
\end{equation}
With this factorisation at hand, we now consider the difference between the NCE objective: $\JN(\thetab)$ in \eqref{eq:J} and the VNCE objective: $\JV(\thetab, q)$ in \eqref{eq:vnce objective}. Each objective consists of two terms: the first is an expectation with respect to the data, the second an expectation with respect to the noise distribution $\pnoise$. The second terms of $\JN$ and $\JV$ are identical, so their difference equals the difference between their first terms
\begin{align}
    & \JN(\thetab) - \JV(\thetab, q)  \nonumber \\
    &= \Ex \log \left( \frac{\pnn_{\thetab}(\x)}{\pnn_{\thetab}(\x) + \nu \pnoise(\x)} \right) 
    -  \Ex \Evar{\x} \log \left( \frac{\pnn_{\thetab}(\x, \z)}{\pnn_{\thetab}(\x, \z) 
    + \nu \pnoise(\x)q(\z \given \x)} \right) \\
    &= \Ex \Evar{\x} \biggl[ \log \left( \frac{\pnn_{\thetab}(\x)}{\pnn_{\thetab}(\x) + \nu \pnoise(\x)} \right) 
    +  \log \left( 1 + \frac{ \nu \pnoise(\x)q(\z \given \x)}{\pnn_{\thetab}(\x) p_{\thetab}(\z \given \x)} \right) \biggr] \\
    &= \Ex \Evar{\x} \left[ \log \left( \frac{\pnn_{\thetab}(\x)}{\pnn_{\thetab}(\x) + \nu \pnoise(\x)} + \frac{ \pnn_{\thetab}(\x)}{\pnn_{\thetab}(\x) + \nu \pnoise(\x)} \frac{ \nu \pnoise(\x)}{\pnn_{\thetab}(\x)} \frac{q(\z \given \x)}{p_{\thetab}(\z \given \x)} \right) \right] \\
    &= \Ex \Evar{\x} \left[ \log \left( \frac{\pnn_{\thetab}(\x)}{\pnn_{\thetab}(\x) + \nu \pnoise(\x)} + \frac{ \nu \pnoise(\x)}{\pnn_{\thetab}(\x) + \nu \pnoise(\x)} \frac{q(\z \given \x)}{p_{\thetab}(\z \given \x)} \right) \right] \\
    &= \Ex \Evar{\x} \left[ \log \left( \frac{\pnn_{\thetab}(\x)}{\pnn_{\thetab}(\x) + \nu \pnoise(\x)} + \left( 1 - \frac{\pnn_{\thetab}(\x)}{\pnn_{\thetab}(\x) + \nu \pnoise(\x)} \right) \frac{q(\z \given \x)}{p_{\thetab}(\z \given \x)} \right) \right] \\
    \label{eq: f divergence formula}
    &= \Ex \Evar{\x} \left[ \log \left( \classprob + (1 - \classprob) \frac{q(\z \given \x)}{p_{\thetab}(\z \given \x)} \right) \right] \\
    &=  \Ex \left[ \infdiv{f_{\x}}{p_{\thetab}(\z \given \x)}{q(\z \given \x)} \right],
\end{align}
where $f_{\x}(u) = \log(\classprob + (1 - \classprob)u^{-1})$. To ensure that that $D_{f_{\x}}$ is a valid f-divergence, we need to prove that $f$ is convex and $f_{\x}(1) = 0$. The latter is trivial, since $f_{\x}(1) = \log(\classprob + (1 - \classprob)) = \log(1) = 0$, and convexity follows directly from Supplementary Materials \ref{appendix:convexity result}.

We now prove that this f-divergence can be expressed as the difference of two KL-divergences as in \eqref{eq: f-divergence is diff of two KLs} in the main text. To do this, we pull $q/p$ outside of the log in \eqref{eq: f divergence formula},
\begin{align}
    &\infdiv{f_{\x}}{p_{\thetab}(\z \given \x)}{q(\z \given \x)} \nonumber \\
    &= \Evar{\x} \left[ \log \frac{q(\z \given \x)}{p_{\thetab}(\z \given \x)} \right] 
    + \Evar{\x} \left[ \log \left( \classprob\frac{p_{\thetab}(\z \given \x)}{q(\z \given \x)} + (1 - \classprob) \right) \right] \\
    &= \Evar{\x} \left[ \log \frac{q(\z \given \x)}{p_{\thetab}(\z \given \x)} \right] 
    - \Evar{\x} \left[ \log \left( \frac{q(\z \given \x)}{\classprob p_{\thetab}(\z \given \x) + (1 - \classprob)q(\z \given \x)} \right) \right] \\
    &= \infdiv{KL}{q(\z \given \x)}{p_{\thetab}(\z \given \x)} - \infdiv{KL}{q(\z \given \x)}{m_{\thetab}(\z, \x)}.
\end{align}
where $m_{\thetab}(\z, \x) = \classprob p_{\thetab}(\z \given \x) + (1 - \classprob)q(\z \given \x)$.

\section{Proof of Theorem \ref{theorem: optimal q}}
\label{appendix: proof of theorem 1}
We first show that
\begin{equation}
        \JN(\thetab) = \JV(\thetab, q) \hspace{2mm} \Leftrightarrow \hspace{2mm} q(\z \given \x) = p_{\thetab}(\z \given \x). 
\end{equation}
We could obtain this result directly from the lower bound in Section \ref{sec:nce lower bound} in the main text. However, for brevity, we make use of the Lemma \ref{lemma: diff J - J1}, where we obtained the equality
\begin{equation}
\JN(\thetab) - \JV(\thetab, q) = \Ex \left[ \infdiv{f_{\x}}{p_{\thetab}(\z \given \x)}{q(\z \given \x)} \right] .
\end{equation} 
The f-divergence on the right-hand side is non-negative and equal to zero if and only if the two posteriors coincide. Hence, $\JN(\thetab) = \JV(\thetab, q)$ if and only if $q(\z \given \x) = p_{\thetab}(\z \given \x)$.

We now show that
\begin{align}
     \infdiv{f_{\x}}{p_{\thetab}(\z \given \x)}{q(\z \given \x)} \ \rightarrow \ \infdiv{KL}{q(\z \given \x)}{p_{\thetab}(\z \given \x)}
 \end{align}
as $\classprob = \pnn_{\thetab}(\x) / (\pnn_{\thetab}(\x) + \nu \pnoise(\x))  \rightarrow 0$.
Again, this follows quickly from Lemma \ref{lemma: diff J - J1}. Specifically, in \eqref{eq: f divergence formula}, we obtained
\begin{equation}
	\JN(\thetab) - \JV(\thetab, q) = \Evar{\x} \left[ \log \left( \classprob + (1 - \classprob) \frac{q(\z \given \x)}{p_{\thetab}(\z \given \x)} \right) \right] .
\end{equation}
As $\classprob \rightarrow 0$, we obtain the standard KL-divergence.

\section{Proof of Theorem \ref{theorem: equivalece of vnce and nce}}
\label{appendix:proof of theorem 2}
Our goal is to show that
\begin{equation}
    \max_{\thetab} \JN(\thetab) = \max_{\thetab} \max_{q} \JV(\thetab, q).
\end{equation}
We know from Theorem \ref{theorem: optimal q} that:
\begin{equation}
    p_{\thetab}(\z \given \x) = \argmax{q} \JV(\thetab, q) ,
\end{equation}
and that, plugging this optimal $q$ into $\JV$ makes the variational lower bound tight,
\begin{equation}
    \JV(\thetab, p_{\thetab}(\z \given \x)) = \JN(\thetab).
\end{equation}
Hence,
\begin{equation}
    \max_{\thetab} \max_{q} \JV(\thetab, q) = \max_{\thetab} \JV(\thetab, p_{\thetab}(\z \given \x)) =  \max_{\thetab} \JN(\thetab) .
\end{equation}

\section{Proof of Corollary \ref{corollary:em for vnce}}
Let $k \in \mathbb{N}$. After the E-step of optimisation, we have $q_k(\z \given \x) = p(\z \given \x ; \thetab_k)$ and so, by Lemma \ref{lemma: diff J - J1},
    \begin{align}
         \JN(\thetab_k) - \JV(\thetab_k, q_k)
        = \Ex \left[ \infdiv{f_{\x}}{p(\z \given \x; \thetab_k)}{p(\z \given \x; \thetab_k)} \right]
        = 0,
    \end{align}
    implying that $\JV(\thetab_k, q_k) = \JN(\thetab_k)$. Now, in the M-step of optimisation, we have
    \begin{equation}
        \thetab_{k+1} = \argmax{\thetab} \JV(\thetab, q_k) \ \ \Longrightarrow \ \  \JV(\thetab_{k+1}, q_k) \geq \JV(\thetab_k, q_k) \ , 
    \end{equation}
    finally, by using Lemma \ref{lemma: diff J - J1} again, we see that $\JN(\thetab_{k+1}) \geq \JV(\thetab_{k+1}, q_k)$. Putting everything together,
    \begin{equation}
        \JN(\thetab_{k+1}) \geq \JV(\thetab_{k+1}, q_k)  \geq \JV(\thetab_k, q_k) = \JN(\thetab_k) \ .
    \end{equation}

\section{Optimal proposal distribution in the second term of the VNCE objective}
\label{appendix:optimal proposal}
We know from Theorem \ref{theorem: optimal q} that the optimal variational distribution is the true posterior, $q(\z \given \y) = \pnorm(\z \given \y; \thetab)$. Thus, we simply need to show that the true posterior is the optimal proposal distribution for the importance sampling (IS) estimate in the second term of the VNCE objective. 

As shown in Supplementary Materials \ref{appendix:proof of lemma}, the following factorisation holds
\begin{equation}
    \pnn_{\thetab}(\y, \z) = \pnn_{\thetab}(\y) p_{\thetab}(\z \given \y) .
\end{equation}
Using this factorisation of $\pnn$, we get
\begin{align}
 \pnn(\y; \thetab) &= \E_{\z \sim q(\z \given \y)} \left[\frac{\pnn(\y,\z; \thetab)}{q(\z \given \y)} \right] 
 \label{eq:vnce second term importance sampling} \\ 
 &= \pnn(\y; \thetab) \E_{\z \sim q(\z \given \y)} \left[\frac{\pnorm(\z \given \y; \thetab)}{q(\z \given \y)} \right] .
 \label{eq: expectation of true and approx ratio}
\end{align}
Hence, the variance of a Monte Carlo estimate of the expectation in \eqref{eq:vnce second term importance sampling} will equal the variance of a Monte Carlo estimate of the expectation in \eqref{eq: expectation of true and approx ratio}. When $q(\z \given \y) = \pnorm(\z \given \y; \thetab)$, the latter expectation equals one, yielding a zero-variance---and thus optimal---Monte Carlo estimate. 

We have therefore shown that the use of IS is optimal when we have access to $\pnorm(\z \given \y; \thetab)$. More generally, it will still be sensible when we have access to a parameterised approximate posterior $q(\z \given \y; \alphab)$, which is close to the true posterior. However, one potential issue that could arise in practice is that $q$ is only close to the true posterior when conditioning on data $\x$, but not when conditioning on noise samples $\y$. This is because we only optimise the parameters of $q$ with respect to the first term of the VNCE objective, in which we only condition on data $\x$. In our experiments, we did not observe such an issue. However, we expect that if $\z$ is high-dimensional and the noise distribution is sufficiently different from the data distribution, then this could become an issue.

\section{Experimental settings for toy approximate inference problem}
\label{sec:Appendix experiment visualising posterior}
In Section \ref{sec: toy model for approx inference} we approximated a posterior $\pnorm(\z \given \x)$ with a variational distribution $q(\z \given \x ; \alphab) = \N(\z ; \ \mub(\x; \alphab), \Sigmab(\x; \alphab))$, where $\Sigmab$ is a diagonal covariance matrix, and $\mub$ and $\Sigmab$ are parametrised by a single 2-layer feed-forward neural network with weights $\alphab$.

The output layer of the neural network has 4 dimensions, containing the concatenated vectors $\mub$ and $\log(\mathrm{diag}(\Sigmab))$. The input to the network is a 2 dimensional vector $\x$ of observed data. In each hidden layer there are 100 hidden units, generated by an affine mapping composed with a $\tanh$ non-linearity applied to the previous layer. The weights of the network are initialised from $\mathcal{U}(-0.05, 0.05)$ and optimised with stochastic gradient ascent in minibatches of $100$ and learning rate of $0.0001$ for a total of $50$ epochs.

\section{Experimental settings for toy parameter estimation (Figure \ref{fig:mse-mog})}
\label{sec:Appendix experiment visualising posterior}
Figure \ref{fig:mse-mog} shows the accuracy of VNCE for parameter estimation using a
population analysis over multiple sample sizes, comparing to NCE and MLE. To produce it, we generated 500 distinct ground-truth values for the standard deviation parameter in the unnormalised MoG, sampling uniformly from the interval $[2, 6]$. For each of the 500 sampled values of $\theta^*$, we estimate $\theta$ using all three estimation methods and with a range of sample sizes. Every run was initialised from five random
values and the best result out of the five was kept in order to avoid
local optima which exist since both the likelihood and NCE objective
functions are bi-modal.

\section{Estimation of noise distribution for undirected graphical model experiments}
Assume the observed data are organised in a matrix $X$ with each column containing all observations of a single variable. We want to fit a univariate truncated Gaussian to each column. To do so, we could estimate the means $\mu_i$ and variances $\sigma_i^2$ of the  \emph{pre-truncated} Gaussians using the following equations \citep{burkardt2014truncated}, where $x_i$ denotes a column of $X$ with empirical mean $\bar{\mu}_i$ and variance $\bar{\sigma}_i^2$:
\begin{align}
    \bar{\mu_i} & =  \mu_i + \frac{\psi(\alpha) }{1 - \Phi(\alpha)} \sigma_i, &
    \bar{\sigma_i^2}  &= \left[ 1 + \frac{\alpha \psi(\alpha) }{1 - \Phi(\alpha)} - \left( \frac{\psi(\alpha) }{1 - \Phi(\alpha)}\right)^2 \right] \sigma_i^2 ,
\end{align}
where $\psi$ is the pdf of a standard normal and $\Phi$ is its cdf. These pairs on non-linear simultaneous equations can then be solved with a variety of methods, such as Newton-Krylov \citep{knoll2004jacobian}. However, whenever $\alpha = \frac{-\mu_i}{\sigma_i} \gg 0$, computing the fractions $\frac{\alpha \psi(\alpha) }{1 - \Phi(\alpha)}$ , $\frac{\psi(\alpha) }{1 - \Phi(\alpha)}$ becomes numerically unstable. In a short note available on GitHub, \citet{Fernandez-de-cossio-Diaz2018} explains how to fix this using the more numerically stable scaled complementary error function $\erfcx(x) = \exp(x^2) \erf(x)$, where $\erf(x)$ is the error function. Introducing the notation
\begin{equation}
    F_1(x) = \frac{1}{\erfcx(x)},  \hspace{10mm} F_2(x) = \frac{x}{\erfcx(x)} ,
\end{equation}
we can then re-express the required fractions in a numerically stable form,
\begin{align}
    \frac{\alpha \psi(\alpha) }{1 - \Phi(\alpha)}  &=  \frac{2}{\sqrt{\pi}} F_2( \frac{\alpha}{\sqrt{2}}), &
    \frac{\psi(\alpha) }{1 - \Phi(\alpha)} &= \frac{2}{\sqrt{\pi}} F_2( \frac{\alpha}{\sqrt{2}}) - \frac{2}{\pi} \left[ F_1( \frac{\alpha}{\sqrt{2}})  \right]^2.
\end{align}

\section{Experimental settings for the undirected graphical model experiments}
For VNCE and NCE we set $\nu=10$, and optimise with
the \texttt{BFGS} optimisation method of Python's \texttt{scipy.optimize.minimize}, capping
the number of iterations at 80. In the case of VNCE, we use
variational-EM, alternating every 5 iterations, and approximating
expectations with respect to the variational distribution with $5$
samples per datapoint. Derivatives with respect to the variational parameters are computed using the reparametrisation trick \citep{kingma2013auto, rezende2014stochastic}, using a standard normal as the base distribution.

For MC-MLE, we apply stochastic gradient ascent
for 80 epochs with minibatches of 100 datapoints. The Monte-Carlo expectations with respect to the posterior distribution and joint distribution use 5 samples per datapoint. These samples are obtained with
the \texttt{tmvtnorm} Gibbs sampler, using the Gibbs sampler from the \texttt{tmvtnorm} package in \texttt{R} with a burnin period of 100 samples and thinning factor of
10.

For VNCE and NCE, we do not enforce positive
semi-definiteness of the matrix $\K$ in \eqref{eq:truncated gaussian},
in line with \citet{lin2016estimation}. For MCMLE, we do enforce it,
since \texttt{tmvtnorm} requires it.

\end{appendix}	

\end{document}